\documentclass[11pt]{article}
\usepackage{acl2001,times,hyphen}
\usepackage{times,epsfig,boxedminipage,url}

\title{The OLAC Metadata Set and Controlled Vocabularies}
\author{Steven Bird \\
  Linguistic Data Consortium \\
  University of Pennsylvania \\
  3615 Market Street, Suite 200 \\ 
  Philadelphia, PA 19104-2608, USA \\ 
  {\tt sb@ldc.upenn.edu} \And
Gary Simons \\
  SIL International \\
  7500 West Camp Wisdom Road \\
  Dallas, TX 75236, USA \\
  {\tt Gary\_Simons@sil.org}}

\date{}

\def\myurl#1{{[\small\url{#1}]}}

\def\elt#1{{\small\sf #1}}
\def\attr#1{{\small\sf #1}}
\def\code#1{{\small\sf #1}}

\begin{document}
\maketitle

\begin{abstract}
As language data and associated technologies proliferate and
as the language resources community rapidly expands,
it has become difficult to locate and reuse existing
resources.  Are there any lexical resources for such-and-such a language?
What tool can work with transcripts in this particular
format?  What is a good format to use for linguistic data of this type?
Questions like these dominate many mailing lists, since web search engines are
an unreliable way to find language resources.
This paper describes a new digital infrastructure for language resource
discovery, based on the Open Archives Initiative, and called
OLAC -- the Open Language Archives Community.
The OLAC Metadata Set and the associated controlled
vocabularies facilitate consistent description and focussed searching.
We report progress on the metadata set and controlled vocabularies, describing
current issues and soliciting input from the language
resources community.
\end{abstract}

\section{Introduction}

Language technology and the linguistic sciences are
confronted with a vast array of \emph{language resources},
richly structured, large and diverse.
Multiple \emph{communities} depend on language resources, including
linguists, engineers, teachers and actual speakers.
Many individuals and institutions provide key pieces of the infrastructure,
including archivists, software developers, and publishers.
Today we have unprecedented opportunities to \emph{connect}
these communities to the language resources they need.
First, inexpensive mass storage technology permits large resources to
be stored in digital form, while
the Extensible Markup Language (XML) and Unicode provide flexible
ways to represent structured data and ensure its long-term survival.
Second, digital publication -- both on and off the world wide web --
is the most practical and efficient means of sharing language resources.
Finally, a standard resource description model, the Dublin Core Metadata
Set, together with an interchange method provided by the Open Archives
Initiative (OAI), make it possible to construct a union catalog over multiple
repositories and archives.

In December 2000, an NSF-funded workshop on Web-Based Language
Documentation and Description, held in Philadelphia, brought together a
group of nearly 100 language software developers, linguists, and archivists
who are responsible for creating language resources in North America, South
America, Europe, Africa, the Middle East, Asia and Australia
\url{http://www.ldc.upenn.edu/exploration/expl2000/}.
The outcome of the workshop was the founding of the
Open Language Archives Community (OLAC),
an application of the OAI to digital archives of
language resources, with the following purpose:

\begin{quote}
OLAC, the Open Language Archives Community, is an international partnership
of institutions and individuals who are creating a worldwide virtual
library of language resources by: (i)~developing consensus on best current
practice for the digital archiving of language resources, and
(ii)~developing a network of interoperating repositories and services for
housing and accessing such resources.
\end{quote}

This paper will describe the leading ideas that motivate OLAC, before
focussing on the metadata set and the controlled vocabularies which
implement part (ii) of OLAC's statement of purpose.
Metadata elements of special interest to the language resources community
include such things as language identification
and language resource type.  The corresponding controlled vocabularies
ensure consistent description.  For example, French language resources
are specified using an official RFC-3066 designation \cite{Alvestrand01},
instead of multiple
distinct text strings like ``French'', ``Francais'' and ``Fran\c{c}ais''.
A separate controlled vocabulary exists for resource type, and has
items such as \code{annotation/phonetic} and \code{description/grammar}.
Services for end-users can map controlled vocabularies onto
convenient terminology for any target language.
(A live demonstration accompanies this presentation.)

\section{Locating Data, Tools and Advice}

We can observe that the
individuals who use and create language resources
are looking for three things: data, tools, and advice.
By DATA we mean any information that documents or describes a language,
such as a published monograph, a computer data file, or
even a shoebox full of hand-written index cards. The information could range
in content from unanalyzed sound recordings to fully transcribed and annotated
texts to a complete descriptive grammar. 
By TOOLS we mean computational resources that facilitate creating, viewing,
querying, or otherwise using language data. Tools include not just software
programs, but also the digital resources that the programs depend on, such as
fonts, stylesheets, and document type definitions.
By ADVICE we mean any information about
what data sources are reliable, what tools are appropriate in a given
situation, what practices to follow when creating new data, and so forth.
In the context of OLAC, the term \emph{language resource} is broadly
construed to include all three of these: data, tools and advice.

\begin{figure}
\centerline{\includegraphics[width=\linewidth]{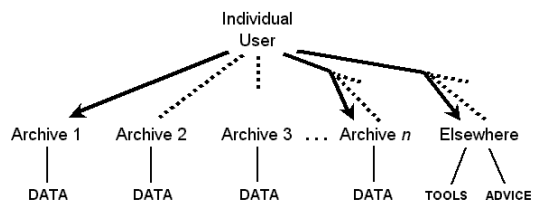}}
\caption{In reality the user can't always get there from here}
\label{fig:vision2}
\end{figure}
Unfortunately, today's user does not have ready access to the resources
that are needed. Figure~\ref{fig:vision2}
offers a diagrammatic view of the reality.
Some archives (e.g. Archive 1) do have a site on the internet which the user is
able to find, so the resources of that archive are accessible. Other archives
(e.g. Archive 2) are on the internet, so the user could access them in theory,
but the user has no idea they exist so they are not accessible in practice.
Still other archives (e.g. Archive 3) are not even on the internet. And there
are potentially hundreds of archives (e.g. Archive $n$) that the user
needs to know about. Tools and advice are out there as well, but are at many
different sites.

There are many other problems inherent
in the current situation. For instance, the user may not be able to find all
the existing data about the language of interest because different sites have
called it by different names (low \emph{recall}).
The user may be swamped with irrelevant resources because search terms
have important meanings in other domains (low \emph{precision}).
The user may not be able to use an accessible
data file for lack of being able to match it with the right tools. The user may
locate advice that seems relevant but have no basis for judging its merits.

\subsection{Bridging the gap}

\subsubsection{Why improved web-indexing is not enough}

As the internet grows and web-indexing technologies improve one might hope
that a general-purpose search engine should be sufficient to bridge the gap
between people and the resources they need, but this is a vain hope.
The first reason is that many language resources, such as audio files
and software, are not text-based.  The second
reason concerns language identification, the single most important
property for describing language resources.  If a language has a canonical name
which is distinctive as a character string, then the user has a chance of
finding any online resources with a search engine.
However, the language may have
multiple names, possibly due to the vagaries of Romanization, such as a
language known variously as Fadicca, Fadicha, Fedija, Fadija, Fiadidja,
Fiyadikkya, and Fedicca (giving low recall).
The language name may collide with a word which has
other interpretations that are vastly more frequent, e.g.\ the language
names Mango and Santa Cruz (giving low precision).

The third reason why general-purpose search engines are inadequate is
the simple fact that much of the material is not,
and will not, be documented in free prose on the web.
Either people will build systematic catalogues of their resources,
or they won't do it at all.
Of course, one can always export a back-end database
as HTML and let the search engines index the materials.
Indeed, encouraging people to document resources and make them
accessible to search engines is part of our vision.
However, despite the power of web search engines, there remain many
instances where people still prefer to use more formal databases to
house their data.

This last point bears further consideration.  The challenge is to
build a system for ``bringing like things together and differentiating among
them'' \cite{Svenonius00}.
There are two dominant storage
and indexing paradigms, one exemplified by traditional databases and one
exemplified by the web.  In the case of language resources, the metadata is
coherent enough to be stored in a formal database, but sufficiently
distributed and dynamic that it is impractical to maintain it centrally.
Language resources occupy the middle ground between the two paradigms, neither of which
will serve adequately.  A new framework is required that permits the best of
both worlds, namely bottom-up, distributed initiatives, along with consistent,
centralized finding aids.  The Dublin Core (DC) and the
Open Archives Initiative provide the framework we need to ``bridge the gap.''

\subsubsection{The Dublin Core Metadata Initiative}

The Dublin Core Metadata Initiative began in 1995 to develop
conventions for resource discovery on the web
\myurl{dublincore.org}.
The Dublin Core metadata elements represent a broad, interdisciplinary
consensus about
the core set of elements that are likely to be widely useful to support
resource discovery.  The Dublin Core consists of 15 metadata elements,
where each element is optional and repeatable: \elt{Title, Creator, Subject,
Description, Publisher, Contributor, Date, Type, Format, Identifier, Source,
Language, Relation, Coverage, Rights}.
This set can be used to describe resources that
exist in digital or traditional formats.

In ``Dublin Core Qualifiers'' \cite{DCQ00}
two kinds of qualifications are allowed: encoding schemes and refinements. An
{\it encoding scheme} specifies a particular controlled vocabulary or notation
for expressing the value of an element. The encoding scheme serves to aid a
client system in interpreting the exact meaning of the element content. A
{\it refinement} makes the meaning of the element more specific.
For example,
a \elt{Language} element can be {\it encoded}
using the conventions of RFC 3066 to unambiguously identify the language
in which the resource is written (or spoken).
A \elt{Subject} element can be given a language {\it refinement}
to restrict its interpretation to concern the language the resource is about.

\subsubsection{The Open Archives Initiative}

The Open Archives Initiative (OAI)
was launched in October 1999 to provide a common framework across
electronic preprint archives, and it has since been broadened
to include digital repositories of scholarly materials regardless
of their type
\myurl{www.openarchives.org} \cite{LagozeVandeSompel01}.

\begin{figure}
\centerline{\includegraphics[width=\linewidth]{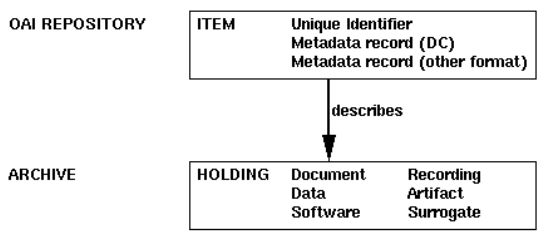}}
\caption{Bridging the gap through community infrastructure}
\label{fig:white-paper1}
\end{figure}
In the OAI infrastructure, each participating archive implements a
repository -- a network accessible server offering public access
to archive holdings. The primary object in an OAI-conformant
repository is called an {\it item}, having a unique identifier
and being associated with one or more metadata records.
Each metadata record describes an archive holding, which is any
kind of primary resource such as a document, raw data, software, a
recording, a physical artifact, a digital surrogate, and so forth.
Each metadata record will usually contain a reference to an entry
point for the holding, such as a URL or a physical location,
as shown in Figure~\ref{fig:white-paper1}.

To implement the OAI infrastructure, a participating archive must comply
with two standards: the {\it OAI shared metadata set} (Dublin Core), which
facilitates interoperability across all repositories participating in the
OAI, and the {\it OAI metadata harvesting protocol}, which allows
software services to query a repository using HTTP requests.

\begin{figure}
\centerline{\includegraphics[width=\linewidth]{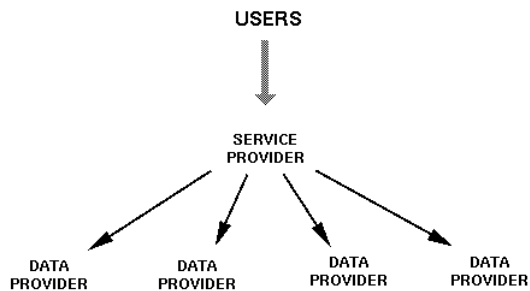}}
\caption{A Service Provider Accessing Multiple Data Providers}
\label{fig:white-paper2}
\end{figure}

OAI archives are called ``data providers,'' though they are strictly just
{\it metadata} providers. Typically, data providers will also have a
submission procedure, together with a long-term storage system, and a
mechanism permitting users to obtain materials from the archive. An OAI
``service provider'' is a third party that provides end-user services (such
as search functions over union catalogs) based on metadata harvested from
one or
more OAI data providers.  Figure~\ref{fig:white-paper2}
illustrates a
single service provider accessing three data providers
(using the OAI metadata harvesting protocol).
End-users only interact with service providers.

Over the past decade, the Linguist List has become the primary
source of online information for
the linguistics community, reaching out to over 13,000
subscribers worldwide, and having four complete mirror sites.
The Linguist List will be augmenting its service by hosting the
primary service provider for OLAC, and permitting end-users to browse
distributed language resources at a single place.

\subsection{Applying the OAI to language resources}

The OAI infrastructure is a new invention;
it has the bottom-up, distributed character of the web,
while simultaneously having the efficient, structured
nature of a centralized database.  This combination is well-suited to
the language resource community, where the available data is growing
rapidly and where a large user-base is fairly consistent in how it describes
its resource needs.

The primary outcome of the Philadelphia
workshop was the founding of the Open Language
Archives Community, and with it the identification of an advisory board, alpha
testers and member archives.  Details of these groups are available from
the OLAC site \myurl{www.language-archives.org}.

Recall that the OAI community is defined by the archives which
comply with the OAI metadata harvesting protocol
and that register with the OAI.
Any compliant repository can register as an Open Archive, and
the metadata provided by an Open Archive is open to the public.
OAI data providers may support metadata standards in addition to the
Dublin Core.  Thus, a specialist community can define a metadata format which is
specific to its domain.  Service providers, data providers and users that
employ this specialized metadata format constitute an OAI \emph{subcommunity}.
The workshop participants agreed unanimously that the
OAI provides a significant piece of the infrastructure
needed for the language resources community.

In the same way that OLAC represents a specialized
subcommunity with respect to the entire Open Archives community, there are
specialized subcommunities within the scope of OLAC.  For
instance, the ISLE Meta Data Initiative is developing a detailed metadata
scheme for corpora of recorded speech events and their associated descriptions
\cite{IMDI00}.
Similarly, the language data centers -- the Linguistic Data Consortium (LDC)
and the European Language Resources Association (ELRA) -- are using OLAC
metadata as the basis of a joint catalog, and will add elements and
vocabularies for their specialized needs (price, rights, and categories
of membership and use).
For archived language resources that are of this kind, such a metadata scheme would
support a richer description.  This specialized subcommunity can implement its own
service provider that offers focused searching based on its own rich metadata
set.  At the same time, the data providers will exposing OLAC and
Dublin Core versions of the metadata, permitting the resources to be
discovered by users of OLAC and OAI service providers.

\subsection{Federation and integration of language resource archives}

\begin{figure*}[tbhp]
\begin{center}
{\normalsize
\framebox{
\begin{tabular}{lp{0.68\textwidth}}
\multicolumn{2}{l}{\normalsize\bf oai:ldc:LDC94T5} \\
Date: 
& 1994\\
Title: 
& ECI Multilingual Text\\
Type:
& text\\
Identifier: 
& 1-58563-033-3\\
Subject.language:
& Albanian, {\bf Bulgarian}, Chinese, Czech, Dutch,
  English, Estonian, French, Gaelic, German, Greek,
  Italian, Japanese, Latin, Lithuanian, Malay,
  Spanish, Danish, Uzbek, Norwegian, Portuguese,
  Russian, Serbian, Swedish, Turkish, Tibetan \\
Identifier:
& http://www.ldc.upenn.edu/Catalog/LDC94T5.html \\
Description: 
& Recommended Applications:
information retrieval, machine translation, language modeling\\[1ex]

\multicolumn{2}{l}{\normalsize\bf oai:elra:L0030} \\
Title:
& Bulgarian Morphological Dictionary \\
Date: 
& 1998 \\
Subject.language:
& {\bf Bulgarian} \\
Description: 
& 67,500 entries divided into 242 inflectional types
(including proper nouns), morphosyntactic information for each
entry, and a morphological engine (MS DOS and WINDOWS 95/NT) for
morphological analysis and generation \\
Identifier:
& http://www.icp.inpg.fr/ELRA/cata/text\_det.html\#bulmodic \\[1ex]

\multicolumn{2}{l}{\normalsize\bf oai:dfki:KPML}\\
Title:
& KPML \\
Creator: 
& Bateman and many others \\
Subject.language: 
& Spanish, Russian, Japanese, Greek, German, French, English, Czech, {\bf Bulgarian}\\
Format.os:
& Windows NT, Windows 98, Windows 95/98, Solaris \\
Type.functionality: 
& Software: Annotation Tools, Grammars, Lexica, Development Tools,
  Formalisms, Theories, Deep Generation, Morphological Generation,
  Shallow Generation \\
Description: 
& Natural Language Generation Linguistic Resource Development and
Maintenance workbench for large scale generation grammar development,
teaching, and experimental generation. Based on systemic-functional
linguistics. Descendent of the Penman NLG system. \\
Identifier:
& http://www.purl.org/net/kpml \\
Description: 
& Contact: bateman@uni-bremen.de \\
Relation.requires:
& Windows: none; Solaris: CommonLisp + CLIM
\end{tabular}}}
\end{center}
\caption{Querying the Prototype Service Provider for Bulgarian Resources}
\label{fig:sp}
\end{figure*}

The OAI framework permits archives to interoperate.  OAI archives support
the Dublin Core metadata format and metadata harvesting protocol.  OLAC
archives additionally support the OLAC metadata format.  Widespread
adoption of these standards will permit language resource archives to
be federated and integrated.

First, a collection of archives which support the same metadata format can be
federated, in the sense that a virtual meta-archive can collect all the
information into a single place, and end-users can query multiple archives
simultaneously.  To demonstrate this,
the Linguistic Data Consortium has harvested the catalogs of
three language resource
archives (LDC, ELRA, DFKI) and created a prototype service provider.
A search for \attr{language=Bulgarian} returns records from all three archives,
as shown in Figure~\ref{fig:sp} \cite{BanikBird01}.

Second, a collection of archives which support the same metadata format can be
integrated, in the sense that relational joins can be performed
across different archives.  This permits queries such as:
``find all lexicon tools that understand a format for which Hungarian
data is available.''

\section{A Core Metadata Set for Language Resources}
\label{sec:metadata}

The OLAC Metadata Set extends the Dublin Core set only to
the minimum degree required to express basic properties
of language resources which are useful as finding aids.

All fifteen Dublin Core elements are used in the OLAC Metadata Set. In
order to suit the specific needs of the language resources community, the
elements have been qualified following principles articulated in
``Dublin Core Qualifiers'' \cite{DCQ00}
and exemplified in \cite{DCQHTML00}.

This section describes some of
the attributes, elements and controlled vocabularies of
the OLAC Metadata Set.  Before launching into this discussion, we first
review some XML terminology and explain some aspects of the OLAC
representation which follow directly from our choice of XML.

\subsection{Aside: XML representation}

The Extensible Markup Language (XML) is the universal format for structured
documents and data on the Web \myurl{www.w3.org/XML}.
The key building block of an XML document is the \emph{element}.
An element has a \emph{name}, \emph{attributes} and \emph{content}.
Here is an example of an element \elt{Language} with attributes
\attr{refine} and \attr{code}, and free-text content:

{\small
\begin{verbatim}
<Language refine="OLAC" code="x-sil-BAN">
  Foreke Dschang</Language>
\end{verbatim}
}

In general, XML elements may contain other elements, or they may be empty.
XML Document Type Definitions (DTDs) and XML schemas are grammars that
define the structure of a valid XML document,
and they limit the arrangement of XML elements in a
document.  We believe it is important to use a formal mechanism for validating
a metadata record.  Following the OAI, we use XML schemas to specify the OLAC
metadata format.

XML schemas make it possible for element content and attribute values
to be constrained according to the element name.  However, XML schemas do not
permit element content to be constrained on the basis of the attribute value.
Accordingly, in implementing qualified Dublin Core using XML,
we are limited to using
one encoding scheme (or controlled vocabulary) per element.

There are two cases we need to consider here.  In the case where all
refinements of an element employ the same encoding scheme, we use the element
name as is and add a \attr{refine} attribute with a fixed value.  This
documents that the particular encoding scheme has been used, and ensures that
the element cannot be confused with a corresponding unqualified
Dublin Core element (see the above example).
In the case where different refinements of an element employ different encoding
schemes, then a unique element must be defined.  Following
\cite{DCQHTML00}, we define such elements by concatenating the
Dublin Core element name
and the refinement name with an intervening dot.  An example is shown below:

{\small
\begin{verbatim}
<Format.encoding code="iso-8859-1"/>
\end{verbatim}
}

\subsection{Attributes used in implementing the OLAC Metadata Set}

Three attributes -- \attr{refine}, \attr{code}, and \attr{lang} -- are used
throughout the metadata set to handle most qualifications to Dublin Core. Some
elements in the OLAC Metadata Set use the \attr{refine} attribute to identify
element refinements. These qualifiers make the meaning of an element narrower
or more specific. A refined element shares the meaning of the unqualified
element, but with a more restricted scope \cite{DCQ00}.

Some elements in the OLAC Metadata Set use the \attr{code} attribute to
hold metadata values that are taken from a specific encoding scheme. When an
element may take this attribute, the attribute value specifies a precise value
for the element taken from a controlled vocabulary or formal notation
(\S\ref{sec:cv}).
In such cases, the element content may also be used
to specify a freeform elaboration of the coded value.

Every element in the OLAC Metadata Set may use the \attr{lang} attribute.
It specifies the language in which the text in the content of the element is
written. The value for the attribute comes from a controlled vocabulary
OLAC-Language.
By default, the \attr{lang} attribute has
the value ``en'', for English. Whenever the language of the element content is
other than English, the \attr{lang} attribute should be used to identify the
language. By using multiple instances of the metadata elements tagged for
different languages, data providers may offer their metadata records in
multiple languages.

In addition, there is a \attr{lang} attribute on the \verb|<olac>|
element that contains the metadata elements for a given metadata record. It
lists the languages in which the metadata record is designed to be read. This
attribute holds a space-delimited list of language codes.
By default, this attribute has
the value ``en'', for English, indicating that the record is aimed only at
English readers. If an explicit value is given for the attribute, then the
record is aimed at readers of all the languages listed.

Service providers should use this information in order to offer
multilingual views of the metadata. When a metadata record lists only one
alternative language, then all elements are displayed (regardless of their
individual languages), unless the user has requested to suppress all records in
that language. When a metadata record has multiple alternative languages, the
user should be able to select one and have display of elements in the other
languages suppressed. An element in a language not included in the list of
alternatives should always be displayed (for instance, the vernacular title of
a work).

\subsection{The elements of the OLAC Metadata Set}

In this section we present a synopsis of the elements of the OLAC metadata
set.  For each element, we provide a one sentence definition followed by a
brief discussion, systematically borrowing and adapting the definitions
provided by the Dublin Core Metadata Initiative \cite{DCMES99}.  Each element
is optional and repeatable.

\begin{description}\setlength{\itemsep}{0pt}\setlength{\parskip}{0pt}
\item[\elt{Contributor}:]
{\bf An entity responsible for making contributions to the content
of the resource.}
Examples of a Contributor include a person, an organization, or a
service.
The \attr{refine} attribute is optionally used to specify the role
played by the named entity in
the creation of the resource, using the controlled vocabulary OLAC-Role.
      
\item[\elt{Coverage}:]
{\bf The extent or scope of the content of the resource.}
Coverage will typically include spatial location or temporal period.
Where the geographical information is predictable from the language identification,
it is not necessary to specify geographic coverage.

\item[\elt{Creator}:]
{\bf An entity primarily responsible for making the content of the resource.}
The \attr{refine} attribute is optionally used to specify the role
played by the named entity in
the creation of the resource, using the controlled vocabulary OLAC-Role.

\item[\elt{Date}:]
{\bf A date associated with an event in the life cycle of the resource.}
The \attr{refine} attribute is optionally used to refine the meaning
of the date using values from a controlled vocabulary (for instance, date of
creation versus date of issue versus date of modification, and so on). The
vocabulary for refinements to Date is defined in \cite{DCQ00}.

\item[\elt{Description}:]
{\bf An account of the content of the resource.}
Description may include but is not limited to: an abstract, table of
contents, reference to a graphical representation of content, or a free-text
account of the content.

\item[\elt{Format}:]
{\bf The physical or digital manifestation of the resource.}
Typically, \elt{Format} may include the media-type or dimensions of the
resource. \elt{Format} may be used to determine the software, hardware or other
equipment needed to use the resource.
The \attr{code} attribute identifies
the format using the controlled vocabulary OLAC-Format.

\item[\elt{Format.cpu}:]
{\bf The CPU required to use a software resource.}
The \attr{code} attribute identifies the CPU using the
controlled vocabulary OLAC-CPU.

\item[\elt{Format.encoding}:]
{\bf An encoded character set used by a digital resource.}
For a digitally encoded text, \elt{Format.encoding} names
the encoded character set it uses. For a font,
\elt{Format.encoding} names an encoded character set that it is able to render. For a
software application, \elt{Format.encoding} names an encoded
character set that it can read or write.
The \attr{code} attribute is used to identify the character set
using the controlled vocabulary OLAC-Encoding.

\item[\elt{Format.markup}:]
{\bf The OAI identifier for the definition of the markup format.}
\elt{Format.markup} provides
an OAI identifier for an XML DTD, schema or some other definition
of the markup format.  (This has the side-effect of ensuring that
the format definition is archived somewhere).
For a software resource,
\elt{Format.markup} names a markup scheme that it can read or write.
The \attr{code} attribute identifies the markup scheme
using the controlled vocabulary OLAC-Markup.

\item[\elt{Format.os}:]
{\bf The operating system required to use a software resource.}
The \attr{code} attribute is used to identify the operating system using the
controlled vocabulary OLAC-OS.  Additional restrictions for
operating system version, may be specified using the element content.

\item[\elt{Format.sourcecode}:]
{\bf The programming language(s) of software distributed in source form.}
The \attr{code} attribute identifies the language using the controlled
vocabulary OLAC-Sourcecode.

\item[\elt{Identifier}:]
{\bf An unambiguous reference to the resource within a given context.}
Recommended best practice is to identify the resource by means of a
string or number conforming to a globally-known formal identification system
(e.g. URIs, ISBNs).
For non-digital archives, Identifier may use
the existing scheme for locating a resource within the collection.

\item[\elt{Language}:]
{\bf A language of the intellectual content of the resource.}
\elt{Language} is used for a language the resource is in, as opposed to the
language it describes (see \elt{Subject.language}).
It identifies a language that the creator of the
resource assumes that its eventual user will understand.
The \attr{code} attribute is used to make a precise
identification of the language using the controlled vocabulary OLAC-Language.

\item[\elt{Publisher}:]
{\bf An entity responsible for making the resource available.}
Examples of a publisher include a person, an organization, or a
service.

\item[\elt{Relation}:]
{\bf A reference to a related resource.}
This element is used to document relationships between resources.
The \attr{refine} attribute is used to refine the nature of the
relationship using values from a controlled vocabulary (for instance, is
replaced by, requires, is part of, and so on). The vocabulary for refinements
to Relation is defined in \cite{DCQ00}.

\item[\elt{Rights}:]
{\bf Information about rights held in and over the resource.}
Typically, a \elt{Rights} element will contain a rights management
statement for the resource, or reference a service providing such information.
Rights information often encompasses intellectual property rights (IPR),
copyright, and various property rights.
The \attr{code} attribute is used to make a summary statement
about rights using the controlled vocabulary OLAC-Rights.

\item[\elt{Rights.software}:]
{\bf Information about rights held in and over a software resource.}
A rights statement pertaining to software, using the controlled
vocabulary OLAC-Software-Rights.

\item[\elt{Source}:]
{\bf A reference to a resource from which the present resource is derived.}
For instance, it
may be the bibliographic information about a printed book of which this is the
electronic encoding or from which the information was extracted.

\item[\elt{Subject}:]
{\bf The topic of the content of the resource.}
Typically, a Subject will be expressed as keywords, key phrases or
classification codes that describe a topic of the resource. Recommended best
practice is to select a value from a controlled vocabulary or formal
classification scheme.

\item[\elt{Subject.language}:]
{\bf A language which the content of the resource describes or discusses.}
As with the Language element, a \attr{code} attribute is
used to identify the language precisely.

\item[\elt{Title}:]
{\bf A name given to the resource.}
Typically, a title will be a name by which the resource is formally known.
A translation of the title can be supplied in a second \elt{Title} element.
The \attr{lang} attribute is used to identify the language of these elements.

\item[\elt{Type}:]
{\bf The nature or genre of the content of the resource.}
The \attr{code} attribute is used to identify the type using the
Dublin Core controlled vocabulary DC-Type.

\item[\elt{Type.data}:]
{\bf The nature or genre of the content of the resource, from a linguistic
standpoint.}
Type includes terms describing general categories, functions, genres,
or aggregation levels for content.
The \attr{code} attribute is used to identify the type using the
controlled vocabulary OLAC-Data.

\item[\elt{Type.functionality}:]
{\bf The functionality of a software resource.}
The \attr{code} attribute is used to identify the type using the
controlled vocabulary OLAC-Functionality.
\end{description}

Observe that some elements, such as \elt{Format}, \elt{Format.encoding}
and \elt{Format.markup}
are applicable to software as well as to data.  Service providers can exploit
this feature to match data with appropriate software tools.

\subsection{The controlled vocabularies}
\label{sec:cv}

Controlled vocabularies are enumerations of legal values for the
\attr{code} attribute.  In some cases, more than one value applies,
in which case the corresponding element must be repeated, once for each
applicable value.  In other cases, no value is applicable ands
the corresponding element is simply omitted.  In yet other cases, the
controlled vocabulary may fail to provide a suitable item, in which case
a similar item can be optionally specified and a prose comment included in the
element content.

\subsubsection{OLAC-Language}

Language identification is an important dimension of language resource
classification. However, the character-string representation of language names
is problematic for several reasons:
different languages (in different parts of the world) may have the
same name;
the same language may have a different name in each country where
it is spoken;
within the same country, the preferred name for a language may
change over time;
in the early history of discovering new languages (before names
were standardized), different people referred to the same language by different
names; and
for languages having non-Roman orthographies, the language name
may have several possible romanizations.
Together, these facts suggest that a standard based
on names will not work.
Instead, we need a standard based on unique identifiers
that do not change, combined with accessible documentation that
clarifies the particular speech variety denoted by each identifier.

The information technology community has a standard for language
identification, namely, ISO 639 \cite{ISO639}. Part 1 of this standard
lists two-letter codes for identifying 160 of the world's major
languages; part 2 of the standard lists three-letter codes for identifying
about 400 languages. ISO 639 in turn forms the core of another standard, RFC
3066 (formerly RFC 1766), which is the
standard used for language identification in the xml:lang attribute of XML and
in the language element of the Dublin Core metadata set.  RFC 3066
provides a mechanism for users to register new language identification codes
for languages not covered by ISO 639, but very few additional languages have
been registered.

Unfortunately, the existing standard falls far short of meeting the
needs of the language resources community since it fails to account for more
than 90\% of the world's languages, and it fails to adequately document what
languages the codes refer to \cite{Simons00}. However, SIL's Ethnologue
\cite{Grimes00} provides a complete system of language identifiers which
is openly available on the Web. OLAC will employ the RFC 3066 extension
mechanism to build additional language identifiers based on the Ethnologue
codes.  For the 130-plus ISO-639-1 codes having a one-to-one mapping onto
Ethnologue codes, OLAC will support both.  Where an ISO code is ambiguous
-- such as \code{mhk} for ``other Mon Khmer languages'' --
OLAC will require the Ethnologue code.
New identifiers for ancient languages, currently being developed by
LINGUIST List, will be incorporated.
These language identifiers are expressed using the \attr{code} attribute of the
\elt{Language} and \elt{Subject.language} elements.
The free-text content of these elements may be used to specify an
alternative human-readable name for the language (where the name
specified by the standard is unacceptable for some reason)
or to specify a dialect (where the resource is dialect-specific).

\subsubsection{OLAC-Data}

After language identification, another dimension of central importance is
the linguistic type of a resource.  Notions such as ``lexicon'' and
``grammar'' are fundamental to OLAC, and the discourse of the
language resources community depends on shared assumptions about what
these types mean.

We believe that it is helpful to distinguish at least four top-level types:
\code{transcription}, \code{annotation}, \code{description} and
\code{lexicon}, each defined broadly as proposed below.
A \code{transcription} is
any time-ordered symbolic representation of a linguistic event.
An \code{annotation} is any kind of structured linguistic information that is
explicitly aligned to some spatial and/or temporal
extent of a linguistic record (such as a recorded signal or an image).
A \code{description} is any description or analysis of a language; unlike a
transcription or an annotation, the structure of a
description is independent of the structure of the
linguistic events that it describes.
A \code{lexicon} is any record-structured inventory of linguistic forms.

For each of these top-level types we envision a more specific vocabulary
to facilitate greater precision.  For example, an orthographic
transcription would have the code \code{transcription/orthographic}.
Other subtypes could include: \code{phonetic}, \code{prosodic},
\code{morphological}, \code{gestural}, \code{part-of-speech},
\code{syntactic}, \code{discourse}, \code{musical}.  The \code{annotation}
type would include these subtypes, and add others
to cover spatial annotation of images (e.g. for OCR annotation of textual
images or for isogloss maps).

The \code{description} type could have subtypes for
\code{grammatical}, \code{phonological}, \code{orthographic}, 
\code{paradigms}, \code{pedagogical}, \code{dialectal} and
\code{comparative}.  The \code{lexicon} type could also carry
subtypes to distinguish wordlists, wordnets, thesauri and
so forth.

\subsubsection{Other controlled vocabularies}

\begin{description}\setlength{\itemsep}{0pt}\setlength{\parskip}{0pt}
\item[OLAC-CPU:]
A vocabulary for identifying the CPU(s) for which the software is
available, in the case of binary distributions:
\code{x86}, \code{mips}, \code{alpha}, \code{ppc}, \code{sparc}, \code{680x0}.

\item[OLAC-Encoding:]
A vocabulary for identifying the character encoding used by a digital
resource, e.g. \code{iso-8859-1}, ...

\begin{figure*}[tb]
{\small\begin{verbatim}
<?xml version="1.0" encoding="UTF-8"?>
<olac
  xmlns="http://www.language-archives.org/OLAC/0.3/"
  xmlns:xsi="http://www.w3.org/2001/XMLSchema-instance"
  xsi:schemaLocation="http://www.language-archives.org/OLAC/0.3/
                http://www.language-archives.org/OLAC/olac-0.3b1.xsd">
  <Title>KPML</Title>
  <Identifier>http://www.purl.org/net/kpml/</Identifier>
  <Creator refine="Author">Bateman, John</Creator>
  <Subject.language code="es"/> <Subject.language code="ru"/>
  <Subject.language code="ja"/> <Subject.language code="el"/>
  <Subject.language code="de"/> <Subject.language code="fr"/>
  <Subject.language code="en"/> <Subject.language code="cs"/>
  <Subject.language code="bg"/>
  <Format.os code="MSWindows/winNT"/> <Format.os code="MSWindows/win95"/>
  <Format.os code="MSWindows/win98"/> <Format.os code="Unix/Solaris"/>
  <Type.functionality>Annotation Tools, Grammars, Lexica, Development Tools,
    Formalisms, Theories, Deep Generation, Morphological Generation,
    Shallow Generation</type.functionality>
  <Relation refine="Requires">Windows: none; Solaris: CommonLisp + CLIM</Relation>
  <Description>Natural Language Generation Linguistic Resource Development and
    Maintenance workbench for large scale generation grammar development,
    teaching, and experimental generation. Based on systemic-functional
    linguistics. Descendent of the Penman NLG system.</Description>
</olac>
\end{verbatim}}
\caption{OLAC Metadata Record for KPML}
\label{fig:olac-record}
\end{figure*}

\item[OLAC-Format:]
A vocabulary for identifying the manifestation of the resource.
The representation is inspired by MIME types, e.g. \code{text/sf} for
SIL standard format.  (\elt{Format.markup} is used to identify the particular
tagset.)  It may be necessary to add new types and subtypes to cover
non-digital holdings, such as manuscripts, microforms, and so forth
and we expect to be able to incorporate an existing vocabulary.

\item[OLAC-Functionality:]
A vocabulary for classifying the functionality of software,
again using the MIME style of representation, and using the
HLT Survey as a source of categories \cite{Cole97} as advocated
by the ACL/DFKI Natural Language Software Registry.  For example,
\code{written/OCR} would cover ``written language input, print or
handwriting optical character recognition.''

\item[OLAC-OS:]
A vocabulary for identifying the operating system(s) for which the software
is available:
\code{Unix}, \code{MacOS}, \code{OS2}, \code{MSDOS}, \code{MSWindows}.
Each of these has optional subtypes, e.g.
\code{Unix/Linux}, \code{MSWindows/winNT}.

\item[OLAC-Rights:]
A vocabulary for classifying the rights held over a resource, e.g.:
\code{open}, \code{restricted}, ...

\item[OLAC-Role:]
A vocabulary for identifying the role of a contributor or creator of the
resource, e.g.: \code{author}, \code{editor}, \code{translator},
\code{transcriber}, \code{sponsor}, ...

\item[OLAC-Software-Rights:]
A vocabulary for classifying the rights held over a resource, e.g.:
\code{open-source}, \code{royalty-free-library},
\code{royalty-free-binary}, \code{commercial}, ...

\item[OLAC-Sourcecode:]
A vocabulary for identifying the programming language(s) used by
software which is distributed in source form, e.g.:
\code{C++}, \code{Java}, \code{Python}, \code{Tcl}, \code{VB}, ...

\end{description}

\section{XML Representation}

The OLAC metadata format consists of an XML schema for the element
set, and a set of schemas for the controlled vocabularies.  The
latest versions are available from the OLAC website.

Figure~\ref{fig:olac-record} shows the OLAC metadata record
corresponding to the KPML display from Figure~\ref{fig:sp}.
The top element is \elt{olac}; this references the XML namespace
for version 0.3b1 of the schema.  The contents of the \elt{olac}
element are the OLAC metadata elements, which are optional and repeatable,
and can occur in any order, as in Dublin Core.

Some elements employ the optional \attr{code} or \attr{refine} attributes,
and/or free-text content.  The third attribute, \attr{lang}, is not used
here since the free-text content is in English (specified in the XML
schema as the default).  For the \elt{Creator} element, the \attr{refine}
attribute narrows the meaning of creator to \code{Author}.  For the
\elt{Subject.language} elements, the \attr{code} attribute specifies
nine languages using Ethnologue codes.  A service provider would map these
codes to human-readable names.

The \elt{Format.os} element illustrates a two-level coding scheme,
consisting of an OS ``family'', followed by a specific operating
system.  Further details can be included in the free-text content if
necessary.  If a piece of software runs on all members of an OS family,
then the more detailed designation can be omitted, e.g. \attr{code="Unix"}.
The \elt{Type.functionality} element is specified using free-text content,
since the details of the controlled vocabulary OLAC-Functionality are still
being worked out.

\section{Conclusions}

The OLAC Metadata Set and controlled vocabularies are works in progress,
and are continuing to be
revised with input from participating archives and members
of the wider language resources community.  We hope to have provided
sufficient motivation and exemplification for our choices so that readers
will easily be able to contribute to ongoing developments.

Even once OLAC is completely in place, there will still be documentation tasks
which the creators of language resources will have to undertake, and new habits to
acquire.  It will always be necessary to identify and manually correct
inconsistent or erroneous metadata.  The OLAC controlled vocabularies will
need to be refined indefinitely in response to changes in the world around us.
The creators of language resources will need to
generate metadata with each new resource and place the resource in a
suitable archive.  The communities will need to adopt best practices for
archival storage formats.

Despite these intrinsic limitations,
the OLAC Metadata Set and controlled vocabularies offer a \emph{template}
for resource description, providing two clear benefits over traditional
full-text description and retrieval.  First, the template guides the
resource creator in giving a \emph{complete description} of the resource,
in contrast to prose descriptions which may omit important details.
And second, the template associates a resource with \emph{standard labels},
such as \elt{creator} and \elt{title}, permitting users to do focussed
searching.
Resources and repositories can proliferate, yet common metadata and
vocabularies will support centralized services giving users easy access to
language resources.

\raggedright\small
\bibliographystyle{acl}

\begin{thebibliography}{}

\bibitem[\protect\citename{Alvestrand}2001]{Alvestrand01}
Harald Alvestrand.
\newblock 2001.
\newblock {RFC} 3066: Tags for the identification of languages (replaces 1766).
\newblock \url{ftp://ftp.isi.edu/in-notes/rfc3066.txt}.

\bibitem[\protect\citename{B\'anik and Bird}2001]{BanikBird01}
\'Eva B\'anik and Steven Bird.
\newblock 2001.
\newblock {LDC} experimental {OLAC} service provider.
\newblock \url{http://wave.ldc.upenn.edu/OLAC/sp-0.2/sp.php4}.

\bibitem[\protect\citename{Cole}1997]{Cole97}
Ronald Cole, editor.
\newblock 1997.
\newblock {\em Survey of the State of the Art in Human Language Technology}.
\newblock Studies in Natural Language Processing. Cambridge University Press.
\newblock \url{http://cslu.cse.ogi.edu/HLTsurvey/}.

\bibitem[\protect\citename{{DCMI}}1999]{DCMES99}
{DCMI}.
\newblock 1999.
\newblock {Dublin Core Metadata Element Set}, version 1.1: Reference
  description.
\newblock \url{http://dublincore.org/documents/1999/07/02/dces/}.

\bibitem[\protect\citename{{DCMI}}2000a]{DCQ00}
{DCMI}.
\newblock 2000a.
\newblock {Dublin Core} qualifiers.
\newblock \url{http://dublincore.org/documents/2000/07/11/dcmes-qualifiers/}.

\bibitem[\protect\citename{{DCMI}}2000b]{DCQHTML00}
{DCMI}.
\newblock 2000b.
\newblock Recording qualified {Dublin Core} metadata in {HTML}.
\newblock \url{http://dublincore.org/documents/2000/08/15/dcq-html/}.

\bibitem[\protect\citename{Grimes}2000]{Grimes00}
Barbara~F. Grimes, editor.
\newblock 2000.
\newblock {\em Ethnologue: Languages of the World}.
\newblock Dallas: Summer Institute of Linguistics, 14th edition.
\newblock \url{http//www.sil.org/ethnologue/}.

\bibitem[\protect\citename{{ISO}}1998]{ISO639}
{ISO}.
\newblock 1998.
\newblock {ISO} 639: Codes for the representation of names of languages-part 2:
  Alpha-3 code.
\newblock \url{http://lcweb.loc.gov/standards/iso639-2/langhome.html}.

\bibitem[\protect\citename{Lagoze and de Sompel}2001]{LagozeVandeSompel01}
Carl Lagoze and Herbert~Van de~Sompel.
\newblock 2001.
\newblock The {Open Archives Initiative}: Building a low-barrier
  interoperability framework.
\newblock \url{http://www.cs.cornell.edu/lagoze/papers/oai-jcdl.pdf}.

\bibitem[\protect\citename{{MPI ISLE Team}}2000]{IMDI00}
{MPI ISLE Team}.
\newblock 2000.
\newblock {ISLE} meta data elements for session descriptions proposal.
\newblock
  \url{http://www.mpi.nl/world/ISLE/documents/draft/ISLE_Metadata_2.0.pdf}.

\bibitem[\protect\citename{Simons}2000]{Simons00}
Gary Simons.
\newblock 2000.
\newblock Language identification in metadata descriptions of language archive
  holdings.
\newblock In Steven Bird and Gary Simons, editors, {\em Proceedings of the
  Workshop on Web-Based Language Documentation and Description}.
\newblock \url{http://www.ldc.upenn.edu/exploration/expl2000/papers/simons/}.

\bibitem[\protect\citename{Svenonius}2000]{Svenonius00}
Elaine Svenonius.
\newblock 2000.
\newblock {\em The Intellectual Foundation of Information Organization}.
\newblock The MIT Press.

\end{thebibliography}

\end{document}